\documentclass[10pt]{article}

\usepackage{latexsym,amsfonts,amsmath,amssymb,mathrsfs,url,amsthm}

\newcommand{\EE}{\mathbb{E}}

\newcommand{\ZZ}{\mathbb{Z}}

\newcommand{\Lcal}{\mathcal{L}}
\newcommand{\Pcal}{\mathcal{P}}
\newcommand{\Zcal}{\mathcal{Z}}

\newcommand{\bsx}{\boldsymbol{x}}
\newcommand{\bsX}{\boldsymbol{X}}
\newcommand{\bsz}{\boldsymbol{z}}
\newcommand{\bsomega}{\boldsymbol{\omega}}
\newcommand{\bstheta}{\boldsymbol{\theta}}
\newcommand{\bsphi}{\boldsymbol{\phi}}

\allowdisplaybreaks

\title{Multilevel Monte Carlo estimation of log marginal likelihood}
\author{Takashi Goda\thanks{School of Engineering, University of Tokyo, Tokyo, Japan ({\tt goda@frcer.t.u-tokyo.ac.jp}; {\tt k.stoneriv@gmail.com})}, Kei Ishikawa\footnotemark[1]}
\date{\today}

\begin{document}
\maketitle
In this short note we provide an unbiased multilevel Monte Carlo estimator of the log marginal likelihood and discuss its application to variational Bayes.

For some dataset $\bsX$ with $N$ i.i.d.\ samples generated from a conditional distribution $p_{\bstheta}(\bsx| \bsz)$ with some random variable $\bsz\sim p_{\theta}(\bsz)$, the log marginal likelihood of $\bsX$ is given by
\[ \log p_{\bstheta}(\bsX) = \sum_{\bsx \in \bsX}\log p_{\bstheta}(\bsx) = N\EE_{\bsX}\left[ \log p_{\bstheta}(\bsx)\right], \]
where, for some importance distribution $q_{\bsphi}(\bsz|\bsx)$ and any positive integer $N_0$, the law of large numbers leads to
\begin{align*}
 \log p_{\bstheta}(\bsx) & = \log \EE_{p_{\bstheta}(\bsz)}\left[p_{\bstheta}(\bsx|\bsz)\right] = \log \EE_{q_{\bsphi}(\bsz|\bsx)}\left[\frac{p_{\bstheta}(\bsx|\bsz)p_{\theta}(\bsz)}{q_{\bsphi}(\bsz|\bsx)}\right] \\
 & = \lim_{\ell\to \infty}\EE_{\bsz_1,\ldots,\bsz_{N_02^{\ell}} \sim q_{\bsphi}(\bsz|\bsx)}\left[\log \frac{1}{N_02^{\ell}}\sum_{i=1}^{N_02^{\ell}}\frac{p_{\bstheta}(\bsx|\bsz_i)p_{\bstheta}(\bsz_i)}{q_{\bsphi}(\bsz_i|\bsx)}\right] .
\end{align*}
In what follows, we write $f_{\bstheta,\bsphi}(\bsx,\bsz) := p_{\bstheta}(\bsx|\bsz)p_{\bstheta}(\bsz)/q_{\bsphi}(\bsz|\bsx).$ Let us introduce a sequence of random variables indexed by $\ell\in \ZZ_{\geq 0}$:
\[ \Pcal_{\bstheta,\bsphi,\ell}(\bsx) = \log \overline{f_{\bstheta,\bsphi}}^{\ell}(\bsx) \quad \text{with $\overline{f_{\bstheta,\bsphi}}^{\ell}(\bsx)=\frac{1}{N_02^{\ell}}\sum_{i=1}^{N_02^{\ell}}f_{\bstheta,\bsphi}(\bsx,\bsz_i)$,}\]
for $\bsz_1,\ldots,\bsz_{N_02^{\ell}} \sim q_{\bsphi}(\bsz|\bsx)$. For the same samples $\bsz_1,\ldots,\bsz_{N_02^{\ell}}$, denote
\[ \Pcal_{\bstheta,\bsphi,\ell-1}^{(a)}(\bsx) = \log \overline{f_{\bstheta,\bsphi}}^{\ell-1,(a)}(\bsx), \quad \Pcal_{\bstheta,\bsphi,\ell-1}^{(b)}(\bsx) = \log \overline{f_{\bstheta,\bsphi}}^{\ell-1,(b)}(\bsx), \]
with
\begin{align*}
 \overline{f_{\bstheta,\bsphi}}^{\ell-1,(a)}(\bsx) & = \frac{1}{N_02^{\ell-1}}\sum_{i=1}^{N_02^{\ell-1}}f_{\bstheta,\bsphi}(\bsx,\bsz_i), \\
 \overline{f_{\bstheta,\bsphi}}^{\ell-1,(b)}(\bsx) & = \frac{1}{N_02^{\ell-1}}\sum_{i=N_02^{\ell-1}+1}^{N_02^{\ell}}f_{\bstheta,\bsphi}(\bsx,\bsz_i),
\end{align*}
and 
\[ \Zcal_{\bstheta,\bsphi,\ell}(\bsx) = \Pcal_{\bstheta,\bsphi,\ell}(\bsx)-\frac{\Pcal_{\bstheta,\bsphi,\ell-1}^{(a)}(\bsx)+\Pcal_{\bstheta,\bsphi,\ell-1}^{(b)}(\bsx)}{2}, \]
for $\ell>0$ and let $\Zcal_{\bstheta,\bsphi,0}(\bsx)\equiv \Pcal_{\bstheta,\bsphi,0}(\bsx)$. Following the idea from multilevel Monte Carlo methods \cite{G08,G15,RG15}, we represent the log marginal likelihood for each data $\bsx\in \bsX$ by a telescoping sum
\[ \log p_{\bstheta}(\bsx) =  \lim_{\ell\to \infty}\EE\left[ \Pcal_{\bstheta,\bsphi,\ell}(\bsx) \right] = \sum_{\ell=0}^{\infty}\EE\left[ \Zcal_{\bstheta,\bsphi,\ell}(\bsx) \right] = \sum_{\ell=0}^{\infty}\omega_\ell \frac{\EE\left[ \Zcal_{\bstheta,\bsphi,\ell}(\bsx) \right]}{\omega_\ell}, \]
for any $\bsomega=(\omega_0,\omega_1,\ldots) $ such that $\omega_\ell>0$ for all $\ell$ and $\|\bsomega \|_1 = 1$, resulting in
\begin{align}\label{eq:teles} \log p_{\bstheta}(\bsX) = N\EE_{\bsX}\left[ \sum_{\ell=0}^{\infty}\omega_\ell \frac{\EE\left[ \Zcal_{\bstheta,\bsphi,\ell}(\bsx) \right]}{\omega_\ell}\right]. 
\end{align}
This representation of the log marginal likelihood naturally leads to an unbiased Monte Carlo estimator
\[ \frac{N}{M}\sum_{m=1}^{M}\frac{\Zcal_{\bstheta,\bsphi,\ell^{(m)}}(\bsx^{(m)})}{\omega_{\ell^{(m)}}}, \]
for any batch size $M>0$, where $\bsx^{(1)},\ldots,\bsx^{(M)}$ are independently and randomly chosen from $\bsX$, whereas $\ell^{(1)},\ldots,\ell^{(M)}\geq 0$ are independently generated from the discrete probability distribution $\bsomega=(\omega_0,\omega_1,\ldots)$. It follows from \cite[Theorem~2 and Remark~3]{GHI19} that the variance of $\Zcal_{\bstheta,\bsphi,\ell}$, denoted by $V_{\ell}$, is of $O(2^{-2\ell})$ if there exist $s>4 $ and $t> 2$ with $(s-4)(t-2)\geq 8$ such that
\[ \EE_{\bsX}\EE_{q_{\bsphi}(\bsz|\bsx)}\left[ \left| \frac{f_{\bstheta,\bsphi}(\bsx,\bsz)}{p_{\bstheta}(\bsx)}\right|^s\right], \EE_{\bsX}\EE_{q_{\bsphi}(\bsz|\bsx)}\left[ \left|\log \frac{f_{\bstheta,\bsphi}(\bsx,\bsz)}{p_{\bstheta}(\bsx)}\right|^t\right]<\infty, \]
where we note that the antithetic property $(\overline{f_{\bstheta,\bsphi}}^{\ell-1,(a)}(\bsx)+\overline{f_{\bstheta,\bsphi}}^{\ell-1,(b)}(\bsx))/2=\overline{f_{\bstheta,\bsphi}}^{\ell}(\bsx)$ is crucial in the proof, which is why we do \emph{not} define $\Zcal_{\bstheta,\bsphi,\ell}$ simply as $\Pcal_{\bstheta,\bsphi,\ell}(\bsx)-\Pcal_{\bstheta,\bsphi,\ell-1}^{(a)}(\bsx)$ (or, $\Pcal_{\bstheta,\bsphi,\ell}(\bsx)-\Pcal_{\bstheta,\bsphi,\ell-1}^{(b)}(\bsx)$) whose variance is as large as $O(2^{-\ell})$. Moreover, since it is obvious that the cost of computing $\Zcal_{\bstheta,\bsphi,\ell}$, denoted by $C_{\ell}$, is of $O(2^{\ell})$, it suffices to choose $\omega_{\ell}\propto 2^{-3\ell/2}$ in order for both the variance $\sum_{\ell=0}^{\infty}V_{\ell}/\omega_{\ell}$ and the expected cost $\sum_{\ell=0}^{\infty}C_{\ell}\omega_{\ell}$ per sample of our estimator to be finite, see \cite[Section~2.2]{G15}.

Let us move on to application of our Monte Carlo estimator to variational Bayes \cite{BKM17}. Instead of maximizing the evidence lower bound 
\begin{align*}
 \Lcal_{\bstheta,\bsphi}(\bsX) & = N \EE_{\bsX}\left[ \log p_{\bstheta}(\bsx)- D_{\mathrm{KL}}\left(q_{\bsphi}(\bsz|\bsx) \parallel p_{\bstheta}(\bsz|\bsx)\right) \right] \\
 & = N \EE_{\bsX}\EE_{q_{\bsphi}(\bsz|\bsx)}\left[ \log f_{\bstheta,\bsphi}(\bsx,\bsz) \right] ,
\end{align*}
with respect to both $\bstheta$ and $\bsphi$ as in \cite{KW14}, our proposal here is to maximize the log marginal likelihood (with respect to $\bstheta$, of course), and at the same time, to maximize the evidence lower bound (or equivalently, to minimize the Kullback-Leibler divergence) with respect to $\bsphi$. Although related works such as \cite{BGS16,N18} have used intermediate quantities between the evidence lower bound and the log marginal likelihood, as far as the authors know, none of them have succeeded in directly looking at the log marginal likelihood when it cannot be evaluated analytically. Using the telescoping sum representation in \eqref{eq:teles}, the gradient of the log marginal likelihood with respect to $\bstheta$ for fixed $\bsphi$ is given by
\[ \nabla_{\bstheta}\log p_{\bstheta}(\bsX) = N\EE_{\bsX}\left[ \sum_{\ell=0}^{\infty}\omega_\ell \frac{\EE\left[\nabla_{\bstheta} \Zcal_{\bstheta,\bsphi,\ell}(\bsx) \right]}{\omega_\ell}\right], \]
where we have
\begin{align*}
\nabla_{\bstheta} \Zcal_{\bstheta,\bsphi,0}(\bsx) & = \frac{\overline{\nabla_{\bstheta} f_{\bstheta,\bsphi}}^{0}(\bsx)}{\overline{f_{\bstheta,\bsphi}}^{0}(\bsx)}, \\
\nabla_{\bstheta} \Zcal_{\bstheta,\bsphi,\ell}(\bsx) & = \frac{\overline{\nabla_{\bstheta} f_{\bstheta,\bsphi}}^{\ell}(\bsx)}{\overline{f_{\bstheta,\bsphi}}^{\ell}(\bsx)} - \frac{1}{2}\left[\frac{\overline{\nabla_{\bstheta} f_{\bstheta,\bsphi}}^{\ell-1,(a)}(\bsx)}{\overline{f_{\bstheta,\bsphi}}^{\ell-1,(a)}(\bsx)}+\frac{\overline{\nabla_{\bstheta} f_{\bstheta,\bsphi}}^{\ell-1,(b)}(\bsx)}{\overline{f_{\bstheta,\bsphi}}^{\ell-1,(b)}(\bsx)}\right],
\end{align*}
for $\ell>0$, whereas the gradient of the evidence lower bound with respect to $\bsphi$ is given by
\[ \nabla_{\bsphi} \Lcal_{\bstheta,\bsphi}(\bsX) = N\EE_{\bsX}\EE_{q_{\bsphi}(\bsz|\bsx)}\left[ \frac{\nabla_{\bsphi} f_{\bstheta,\bsphi}(\bsx,\bsz)}{f_{\bstheta,\bsphi}(\bsx,\bsz)}+\log f_{\bstheta,\bsphi}(\bsx,\bsz) \nabla_{\bsphi}\log q_{\bsphi}(\bsz|\bsx)\right]. \]
This way we can construct unbiased Monte Carlo estimators for both of the gradients $\nabla_{\bstheta}\log p_{\bstheta}(\bsX)$ and $\nabla_{\bsphi} \Lcal_{\bstheta,\bsphi}(\bsX)$, which are
\begin{align*}
& \frac{N}{M}\sum_{m=1}^{M}\frac{\nabla_{\bstheta} \Zcal_{\bstheta,\bsphi,\ell^{(m)}}(\bsx^{(m)})}{\omega_{\ell^{(m)}}} \quad \text{and} \\
& \frac{N}{M}\sum_{m=1}^{M}\overline{\frac{\nabla_{\bsphi} f_{\bstheta,\bsphi}(\bsx^{(m)},\cdot)}{f_{\bstheta,\bsphi}(\bsx^{(m)},\cdot)}+\log f_{\bstheta,\bsphi}(\bsx^{(m)},\cdot) \nabla_{\bsphi}\log q_{\bsphi}(\cdot|\bsx^{(m)})}^{\ell^{(m)}},
\end{align*}
respectively, for any batch size $M>0$, in which the common stochastic samples $\bsx^{(1)},\ldots,\bsx^{(M)}$ and those on $\bsz$ for each $\bsx^{(m)}$ can be used. Finally it must be pointed out that, as inferred from \cite[Lemma~2]{HGGT19} studied in a quite different context, by exploiting the properties $(\overline{f_{\bstheta,\bsphi}}^{\ell-1,(a)}(\bsx)+\overline{f_{\bstheta,\bsphi}}^{\ell-1,(b)}(\bsx))/2=\overline{f_{\bstheta,\bsphi}}^{\ell}(\bsx)$ and $(\overline{\nabla_{\bstheta}f_{\bstheta,\bsphi}}^{\ell-1,(a)}(\bsx)+\overline{\nabla_{\bstheta}f_{\bstheta,\bsphi}}^{\ell-1,(b)}(\bsx))/2=\overline{\nabla_{\bstheta}f_{\bstheta,\bsphi}}^{\ell}(\bsx)$, the variance of every component of $\nabla_{\bstheta} \Zcal_{\bstheta,\bsphi,\ell}(\bsx)$ is shown to be of $O(2^{-2\ell})$ if
\[ \EE_{\bsX}\EE_{q_{\bsphi}(\bsz|\bsx)}\left[ \left| \frac{f_{\bstheta,\bsphi}(\bsx,\bsz)}{p_{\bstheta}(\bsx)}\right|^4\right], \sup_{\bsx,\bsz}\left\|\nabla_{\bstheta}\log f_{\bstheta,\bsphi}(\bsx,\bsz)\right\|_{\infty} <\infty, \]
so that an adequate choice for $\bsomega=(\omega_0,\omega_1,\ldots)$ will remain the same, i.e., $\omega_{\ell}\propto 2^{-3\ell/2}$, even for gradient estimations.

The authors plan to report, in the near future, more applications of multilevel Monte Carlo approaches to various Bayesian computations, such as variational inference for global latent variables using locally marginalized evidence lower bound and computations of various metrics and their gradients including mutual information, reversed KL divergence, variational Renyi's bound and $\chi$-upper bound.


\begin{thebibliography}{20}
\bibitem{BKM17} Blei, D.~M., Kucukelbir, A., McAuliffe, J.~D. (2017) Variational inference: a review for statisticians, Journal of the American Statistical Association, 112 (518), 859--877.
\bibitem{BGS16} Burda, Y., Grosse, R., Salakhutdinov, R. (2016) Importance weighted autoencoders, arXiv:1509.00519.
\bibitem{G08} Giles, M.~B. (2008) Multilevel Monte Carlo path simulation, Operations Research, 56, 607--617.
\bibitem{G15} Giles, M.~B. (2015) Multilevel Monte Carlo methods, Acta Numerica, 24, 259--328.
\bibitem{GHI19} Goda, T., Hironaka, T., Iwamoto, T. (2019) Multilevel Monte Carlo estimation of expected information gains, arXiv:1811.07546 (accepted for publication in Stochastic Analysis and Applications).
\bibitem{HGGT19} Hironaka, T., Giles, M.~B., Goda, T., Thom, H. (2019) Multilevel Monte Carlo estimation of the expected value of sample information, arXiv:1909.00549.
\bibitem{KW14} Kingma, D.~P., Welling, M. (2014) Auto-encoding variational {B}ayes, arXiv:1312.6114.
\bibitem{N18} Nowozin, S. (2018) Debiasing evidence approximations: on importance-weighted autoencoders and Jackknife variational inference, ICLR 2018 conference paper.
\bibitem{RG15} Rhee, C.~H., Glynn, P. (2015) Unbiased estimation with square root convergence for SDE models, Operations Research, 63, 1026--1043.
\end{thebibliography}
\end{document}